\documentclass[11pt]{article}
\usepackage{algorithm2e}
\usepackage{framed}
\usepackage{amssymb}
\usepackage{amsfonts}
\usepackage{amsmath}
\usepackage{graphicx}
\usepackage{subfigure}
\usepackage{url}

\def\a{{\mathbf{\alpha}}}
\def\de{{\mathbf{\delta}}}
\def\De{{{\Delta}}}
\def\l{{\mathbf{\lambda}}}
\def\Nl{{N^{\Lambda}}}

\def\eps{{\epsilon}}

\def\beq{\begin{eqnarray}}
\def\eeq{\end{eqnarray}}
\def\beqs{\begin{eqnarray*}}
\def\eeqs{\end{eqnarray*}}
\newcommand{\fat}{\mathrm{fat}}

\newcommand{\R}{\mathbb{R}}
\newcommand{\RR}{\mathbb{R}}

\newcommand{\x}{\mathbf{x}}

\newcommand{\I}{\mathbb{I}}
\newcommand{\II}{\mathcal{I}}

\newcommand{\EE}{\mathbb{E}}
\newcommand{\FF}{\mathcal{F}}
\newcommand{\E}{\mathbb{E}}
\newcommand{\p}{\mathbb{P}}
\newcommand{\PP}{\mathcal P}
\newcommand{\BB}{\mathcal B}

\newcommand{\HH}{\mathcal H}
\newcommand{\Hnl}{H_{ann}^{\Lambda}}

\newcommand{\ra}{\rightarrow}

\newlength{\defbaselineskip}
\newtheorem{definition}{Definition}
\newtheorem{lemma}{Lemma}
\newtheorem{theorem}{Theorem}

\newenvironment{proof}{\noindent {\em Proof:}}{\\\hspace*{\fill}\mbox{$\diamond$}}

\begin{document}


\title{
Learning with Spectral Kernels and Heavy-Tailed Data%
\footnote{This version of this paper is substantially revised and extended relative to the first version of 24 June 2009.}
}

\author{
Michael W. Mahoney
\thanks{
Department of Mathematics,
Stanford University,
Stanford, CA 94305,
{\tt mmahoney@cs.stanford.edu}.
}
\and
Hariharan Narayanan
\thanks{
Laboratory for Information and Decision Systems,
MIT,
Cambridge, MA 02139,
{\tt har@mit.edu}.
}
}

\date{}
\maketitle

\begin{abstract}
Two ubiquitous aspects of large-scale data analysis are that the data often
have heavy-tailed properties and that diffusion-based or spectral-based
methods are often used to identify and extract structure of interest.
Perhaps surprisingly, popular distribution-independent methods such as those
based on the VC dimension fail to provide nontrivial results for even
simple learning problems such as binary classification in these two
settings.
In this paper, we develop distribution-dependent learning methods that
can be used to provide dimension-independent sample complexity bounds for
the binary classification problem in these two popular settings.
In particular, we provide bounds on the sample complexity of maximum margin 
classifiers when the magnitude of the entries in the feature vector decays
according to a power law and also when learning is performed with the
so-called Diffusion Maps kernel.
Both of these results rely on bounding the annealed entropy of gap-tolerant
classifiers in a Hilbert space.
We provide such a bound, and we demonstrate that our proof technique
generalizes to the case when the margin is measured with respect to more
general Banach space norms.
The latter result is of potential interest in cases where modeling the
relationship between data elements as a dot product in a Hilbert space is
too restrictive.
\end{abstract}



\section{Introduction}
\label{sxn:intro}

Two ubiquitous aspects of large-scale data analysis are that the data often
have heavy-tailed properties and that diffusion-based or spectral-based
methods are often used to identify and extract structure of interest.
In the absence of strong assumptions on the data, 
popular distribution-independent methods such as those based on the VC
dimension fail to provide nontrivial results for even simple learning
problems such as binary classification in these two settings.
At root, the reason is that in both of these situations the data are
formally very high dimensional and that (without additional regularity
assumptions on the data) there may be a small number of ``very outlying''
data points.
In this paper, we develop distribution-dependent learning methods that
can be used to provide dimension-independent sample complexity bounds for
the maximum margin version of the binary classification problem in these 
two popular settings.
In both cases, 
we are able to obtain nearly optimal linear classification hyperplanes since
the distribution-dependent tools we employ are able to control the aggregate
effect of the ``outlying'' data points.
In particular, our results will hold even though the data may be 
infinite-dimensional and unbounded.

\subsection{Overview of the problems}
\label{sxn:intro_overview}

Spectral-based kernels have received a great deal of attention recently in
machine learning for data classification, regression, and exploratory data
analysis via dimensionality reduction~\cite{SWHSL06}.
Consider, for example, Laplacian Eigenmaps~\cite{BN03} and the related
Diffusion Maps~\cite{Lafon06}.
Given a graph $G=(V,E)$ (where this graph could be constructed from the data
represented as feature vectors, as is common in machine learning, or it
could simply be a natural representation of a large social or information
network, as is more common in other areas of data analysis), let
$f_0, f_1, \dots, f_n$ be the eigenfunctions of the normalized Laplacian of
$G$ and let $\l_0, \l_1, \dots, \l_n$ be the corresponding eigenvalues.
Then, the Diffusion Map is the following feature map
$$
\Phi: v \mapsto(\l_0^k f_0(v), \dots, \l_n^k f_n(v))  ,
$$
and Laplacian Eigenmaps is the special case when $k=0$.
In this case, the support of the data distribution is unbounded as the size
of the graph increases; the VC dimension of hyperplane classifiers is
$\Theta(n)$; and thus existing results do not give dimension-independent
sample complexity bounds for classification by Empirical Risk Minimization
(ERM).
Moreover, it is possible (and indeed quite common in certain applications) 
that on some vertices $v$ the eigenfunctions
fluctuate wildly---even on special classes of graphs, such as random graphs
$G(n, p)$, a non-trivial uniform upper bound stronger than $O(n)$ on
$\|\Phi(v)\|$ over all vertices $v$ does not appear to be known.%
\footnote{It should be noted that, while potentially problematic for what we
are discussing in this paper, such eigenvector localization often has a
natural interpretation in terms of the processes generating the data and can
be useful in many data analysis applications.
For example, it might correspond to a high degree node or an articulation
point between two clusters in a large informatics
graph~\cite{LLDM08_communities_CONF,LLDM08_communities_TR,LLM10_communities_CONF};
or it might correspond to DNA single-nucleotide polymorphisms that are
particularly discriminative in simple models that are chosen for
computational rather than statistical reasons~\cite{CUR_PNAS,Paschou07b}.}
Even for maximum margin or so-called ``gap-tolerant'' classifiers, defined 
precisely in Section~\ref{sxn:background} and which are easier to learn
than ordinary linear hyperplane classifiers, the existing bounds of Vapnik
are not independent of the number $n$ of nodes.%
\footnote{VC theory provides an upper bound of
$O\left(\left(n/\Delta\right)^2\right)$ on the VC dimension of gap-tolerant
classifiers applied to the Diffusion Map feature space corresponding to a
graph with $n$ nodes.
(Recall that by Lemma~\ref{lem:vap1} below, the VC dimension of the space of
gap-tolerant classifiers corresponding to a margin $\Delta$, applied to a
ball of radius $R$ is $\sim \left(R/\Delta\right)^2$.)
Of course, although this bound is quadratic in the number of nodes, VC
theory for ordinary linear classifiers gives an $O(n)$ bound.}

A similar problem arises in the seemingly very-different situation that the
data exhibit heavy-tailed or power-law behavior.
Heavy-tailed distributions are probability distributions with tails that are
not exponentially bounded~\cite{Resnick07,CSN09}.
Such distributions can arise via several mechanisms, and they are ubiquitous
in applications~\cite{CSN09}.
For example, graphs in which the degree sequence decays according to a power
law have received a great deal of attention recently.
Relatedly, such diverse phenomenon as the distribution of packet
transmission rates over the internet, the frequency of word use in common
text, the populations of cities, the intensities of earthquakes, and the
sizes of power outages all have heavy-tailed behavior.
Although it is common to normalize or preprocess the data to remove the
extreme variability in order to apply common data analysis and machine
learning algorithms, such extreme variability is a fundamental property of
the data in many of these application domains.

There are a number of ways to formalize the notion of heavy-tailed behavior
for the classification problems we will consider, and in this paper we will
consider the case where the \emph{magnitude} of the entries decays according 
to a power law.
(Note, though, that in Appendix~\ref{sxn:learnHT}, we will, for completeness,
consider the case in which the probability that an entry is nonzero decays in 
a heavy-tailed manner.) 
That is, if
$$
\Phi: v \mapsto(\phi_0(v), \dots, \phi_n(v))
$$
represents the feature map, then $\phi_i(v) \leq Ci^{-\alpha}$ for some
absolute constant $C>0$, with $\alpha > 1$.
As in the case with spectral kernels, in this heavy-tailed situation, the
support of the data distribution is unbounded as the size of the graph
increases, and the VC dimension of hyperplane classifiers is $\Theta(n)$.
Moreover, although there are a small number of ``most important'' features,
they do not ``capture'' most of the ``information'' of the data.
Thus, when calculating the sample complexity for a classification task for
data in which the feature vector has heavy-tailed properties, bounds that do
not take into account the distribution are likely to be very weak.

In this paper, we develop distribution-dependent bounds for problems in
these two settings.
Clearly, these results are of interest since VC-based arguments fail to
provide nontrivial bounds in these two settings, in spite of ubiquity of
data with heavy-tailed properties and the widespread use of spectral-based
kernels in many applications.
More generally, however, these results are of interest since the
distribution-dependent bounds underlying them provide insight into how
better to deal with heterogeneous data with more realistic noise
properties.

\subsection{Summary of our main results}
\label{sxn:intro_summary1}

Our first main result provides bounds on classifying data whose
magnitude decays in a heavy-tailed manner.
In particular, in the following theorem we show that if the weight of the
$i^{th}$ coordinate of random data point is less than $C i^{-\a}$ for some
$C>0, \a > 1$, then the number of samples needed before a maximum-margin classifier is approximately optimal with high probability is independent of the number of features.
\begin{theorem} [Heavy-Tailed Data]
\label{thm:learn_heavytail}
Let the data be heavy-tailed in that the feature vector:
$$
\Phi: v \mapsto(\phi_1(v), \dots, \phi_n(v))  ,
$$
satisfy $|\phi_i(v)| \leq Ci^{-\alpha}$ for some absolute constant $C>0$, with
$\alpha > 1$. Let $\zeta(\cdot)$ denote the Riemann zeta function.
Then, for any $\ell$, if a maximum margin classifier has a margin $>  \De $, with probability more than $1 - \de$, its risk  is less than
$$ \eps :=  \frac{\tilde{O}\left( \frac{\sqrt{\zeta(2 \a) \ell}}{\Delta}\right)+ \log \frac{1}{\de}}{\ell},$$
   where $\tilde{O}$ hides multiplicative polylogarithmic factors.
\end{theorem}
This result follows from a bound on the annealed entropy of gap-tolerant
classifiers in a Hilbert space that is of independent interest.
In addition, it makes important use of the fact that although individual
elements of the heavy-tailed feature vector may be large, the vector has
bounded moments.

Our second main result provides bounds on classifying data with spectral
kernels.
In particular, in the following theorem
we give dimension-independent upper bounds on the sample complexity of
learning a nearly-optimal maximum  margin classifier
in the feature space of the Diffusion Maps.
\begin{theorem} [Spectral Kernels]
\label{thm:learn_spectral}
Let the following Diffusion map be given:
$$
\Phi: v \mapsto(\l_1^k f_1(v), \dots, \l_n^k f_n(v))  ,
$$ where $f_i$ are normalized eigenfunctions (whose $\ell_2(\mu)$) norm is $1$, $\mu$ being the uniform distribution), $\l_i$ are the eigenvalues of the corresponding Markov Chain and $k \geq 0$.
Then, for any $\ell$, if a maximum margin classifier has a margin $>  \De $, with probability more than $1 - \de$, its risk  is less than
$$ \eps :=  \frac{\tilde{O}\left( \frac{\sqrt{\ell}}{\Delta}\right)+ \log \frac{1}{\de}}{\ell},$$
   where $\tilde{O}$ hides multiplicative polylogarithmic factors.
\end{theorem}
As with the proof of our main heavy-tailed learning result, the proof of our
main spectral learning result makes essential use of an upper bound on the
annealed entropy of gap-tolerant classifiers.
In applying it, we make important use of the fact that although individual
elements of the feature vector may fluctuate wildly, the norm of the
Diffusion Map feature vector is bounded.

As a side remark, note that we are not viewing the feature map in
Theorem~\ref{thm:learn_spectral} as necessarily being either a random
variable or requiring knowledge of some marginal distribution---as might
be the case if one is generating points in some space according to some
distribution; then constructing a graph based on nearest neighbors; and
then doing diffusions to construct a feature map.
Instead, we are thinking of a data graph in which the data are adversarially
presented, \emph{e.g.}, a given social network is presented, and diffusions
and/or a feature map is then constructed.

These two theorems provides a dimension-independent (\emph{i.e.},
independent of the size $n$ of the graph and the dimension of the feature
space) upper bound on the number of samples needed to learn a maximum 
margin classifier, under 
the assumption that a heavy-tailed feature map or the Diffusion Map kernel
of some scale is used as the feature map.
As mentioned, both proofs (described below in Sections~\ref{sxn:gapHS_HTapp}
and~\ref{sxn:gapHS_SKapp}) proceed by providing a
dimension-independent upper bound on the annealed entropy of gap-tolerant
classifiers in the relevant feature space, and then appealing to
Theorem~\ref{thm:Vapnik} (in Section~\ref{sxn:background}) relating the
annealed entropy to the generalization error.
For this bound on the annealed entropy of these gap-tolerant classifiers,
we crucially use the fact that
$\E_v \|\Phi(v)\|^2$ is bounded, even if $\sup_v \|\Phi(v)\|$ is
unbounded as~$n \ra \infty$.
That is, although bounds on the individual entries of the feature map do not
appear to be known, we crucially use that there exist nontrivial bounds on
the magnitude of the feature vectors.
Since this bound is of more general interest, we describe it separately.

\subsection{Summary of our main technical contribution}
\label{sxn:intro_summary2}

The distribution-dependent ideas that underlie our two main results
(in Theorems~\ref{thm:learn_heavytail} and~\ref{thm:learn_spectral}) can
also be used to bound the sample complexity of a classification task more
generally under the assumption that the expected value of a norm of the data
is bounded, \emph{i.e.}, when the magnitude of the feature vector of the
data in some norm has a finite moment.
In more detail:
\begin{itemize}
\item
Let $\PP$ be a probability measure on a Hilbert space $\HH$,
and let $\Delta>0$.
In Theorem~\ref{thm:ddb_HS} (in Section~\ref{sxn:gapHS_annent}), we prove
that if $\EE_{\mathcal{P}} \|x\|^{2}  = r^{2} < \infty$,
then the annealed entropy of gap-tolerant classifiers (defined in Section~\ref{sxn:background}) in $\HH$ can be
upper bounded in terms of a function of $r$, $\Delta$, and (the number of
samples) $\ell$, independent of the (possibly infinite) dimension of $\HH$.
\end{itemize}
It should be emphasized that the assumption that the expectation of some
moment of the norm of the feature vector is bounded is a \emph{much} weaker
condition than the more common assumption that the largest element is
bounded, and thus this result is likely of more general interest in dealing
with heterogeneous and noisy data.
For example, similar ideas have been applied recently to the problem of
bounding the sample complexity of learning smooth cuts on a low-dimensional
manifold~\cite{NN09}.

To establish this result, we use a result
(See Lemma~\ref{lem:vap1} in Section~\ref{sxn:gapHS_vcdim}.)
that the VC
dimension of gap-tolerant classifiers in a Hilbert space when the margin is
$\Delta$ over a bounded domain such as a ball of radius $R$ is bounded above
by $\lfloor R^2/\Delta^2 \rfloor + 1$.
Such bounds on the VC dimension of gap-tolerant classifiers have been stated
previously by Vapnik~\cite{Vapnik98}.
However, in the course of his proof bounding the VC dimension of a
gap-tolerant classifier whose margin is $\Delta$ over a ball of radius $R$
(See~\cite{Vapnik98}, page~353.), Vapnik states, without further
justification, that due to symmetry the set of points in a ball that is
extremal in the sense of being the hardest to shatter with gap-tolerant
classifiers is the regular simplex.
Attention has been drawn to this fact by Burges (See~\cite{Burges98},
footnote~20.), who mentions that a rigorous proof of this fact seems to be
absent.
Here, we provide a new proof of the upper bound on the VC dimension of such
classifiers without making this assumption.
(See Lemma~\ref{lem:vap1} in Section~\ref{sxn:gapHS_vcdim} and
its proof.)
Hush and Scovel~\cite{HS01} provide an alternate proof of Vapnik's claim;
it is somewhat different than ours, and they do not extend their proof to 
Banach spaces.

The idea underlying our new proof of this result generalizes to the case
when the data need not have compact support and where the margin may be
measured with respect to more general norms.
In particular, we show that the VC dimension of gap-tolerant classifiers with
margin $\Delta$ in a ball of radius $R$ in a Banach space of Rademacher type
$p \in (1, 2]$ and type constant $T$ is bounded above
by $\sim \left(3TR/\Delta\right)^{\frac{p}{p-1}}$, and that there exists a Banach space of type $p$ (in fact $\ell_p$) for which the VC dimension is  bounded below by $\left(R/\Delta\right)^{\frac{p}{p-1}}$.
(See Lemmas~\ref{lem:banach_ub} and~\ref{lem:banach_lb} in
Section~\ref{sxn:gapBS_vcdim}.)
Using this result, we can also prove bounds for the annealed entropy of
gap-tolerant classifiers in a Banach space.
(See Theorem~\ref{thm:ddb_BS} in Section~\ref{sxn:gapBS_annent}.)
In addition to being of interest from a theoretical perspective, this result
is of potential interest in cases where modeling the relationship between
data elements as a dot product in a Hilbert space is too restrictive, and
thus this may be of interest, \emph{e.g.}, when the data are extremely
sparse and heavy-tailed.

\subsection{Maximum margin classification and ERM with gap-tolerant classifiers}
\label{sxn:intro_maxmargin}

Gap-tolerant classifiers---see Section~\ref{sxn:background} for more
details---are useful, at least theoretically, as a means of implementing
structural risk minimization (see, \emph{e.g.}, Appendix A.2
of~\cite{Burges98}).
With gap-tolerant classifiers, the margin $\De$ is fixed before hand, and
does not depend on the data.
See, \emph{e.g.},~\cite{FS99b,HBS05,HS01,SBWA98}.
With maximum margin classifiers, on the other hand, the margin is a function
of the data.
In spite of this difference, the issues that arise in the analysis of these
two classifiers are similar.
For example, through the fat-shattering dimension, bounds can be obtained
for the maximum margin classifier, as shown by Shawe-Taylor
\emph{et al.}~\cite{SBWA98}.
Here, we briefly sketch how this is achieved.
\begin{definition} Let $\FF$ be a set of real valued functions. We say that a set of points $x_1, \dots, x_s$ is $\gamma-$shattered by $\FF$ if there are real numbers $t_1, \dots, t_s$ such that for all binary vectors $\mathbf{b} = (b_1, \dots, b_s)$  and each $ i \in [s]=\{1,\ldots,s\}$, there is a function $f_{\mathbf{b}}$ satisfying,
\beq f_{\mathbf{b}}(x_i) = \left\{ \begin{array}{ll}
    > t_i + \gamma, & \hbox{if $ b_i = 1$;} \\
    < t_i - \gamma, & \hbox{otherwise.}
  \end{array} \right. \eeq
  For each $\gamma > 0$, the {\it fat shattering dimension} $\fat_\FF(\gamma)$ of the set $\FF$ is defined to be the size of the largest $\gamma-$shattered set if this is finite; otherwise it is declared to be infinity.
  \end{definition}
Note that, in this definition, $t_i$ can be different for different $i$, which is not the case
in gap-tolerant classifiers.
However, one can incorporate this shift into the feature space by a simple
construction.
We start with the following definition of a Banach space of type $p$ with
type constant $T$.

\begin{definition} [Banach space, type, and type constant]
\label{def:rad}
A {\it Banach space} is a complete normed vector space.
A Banach space $\mathcal{B}$ is said to have (Rademacher) type $p$  if there
exists $T < \infty$ such that for all $n$ and $x_1, \dots, x_n \in
\mathcal{B}$
\beqs
\EE_\epsilon[\|\sum_{i=1}^n \epsilon_i x_i\|_{\mathcal{B}}^p]
   \leq T^p \sum_{i=1}^n \|x_i\|_{\mathcal{B}}^p   .
\eeqs
The smallest $T$ for which the above holds with $p$ equal to the type, is
called the type constant of~$\mathcal{B}$.
\end{definition}
Given a Banach space $\BB$ of type $p$ and type constant $T$, let $\BB'$
consist of all tuples $(v, c)$ for $v \in \BB$ and $c \in \RR$, with the
norm $$\|(v, c)\|_{\BB'} := \left(\|v\|^p + |c|^p\right)^{1/p}.$$
Noting that if $\BB$ is a Banach space of type $p$ and type constant $T$
(see Sections~\ref{sxn:gapBS_prelim} and~\ref{sxn:gapBS_vcdim}), one can
easily check that $\BB'$ is a Banach space of type $p$ and type constant
$\max(T, 1)$.

In our distribution-specific setting, we cannot control the fat-shattering
dimension, but we can control the logarithm of the expected value of
$2^{\kappa(\fat_\FF(\gamma))}$ for any constant $\kappa$ by applying
Theorem~\ref{thm:ddb_BS} to $\BB'$.
As seen from Lemma 3.7 and Corollary 3.8 of the journal version
of~\cite{SBWA98}, this is all that is required for obtaining
generalization error bounds for maximum margin classification.
In the present context, the logarithm of the expected value of the
exponential of the fat shattering dimension of linear $1$-Lipschitz
functions on a random data set of size $\ell$ taken i.i.d from $\PP$
on $\BB$ is bounded by the annealed entropy of gap-tolerant classifiers
on $\BB'$ with respect to the push-forward $\PP'$ of the measure $\PP$
under the inclusion $\BB \hookrightarrow \BB'$.

This allows us to state the following theorem, which is an analogue of
Theorem~4.17 of the journal version of~\cite{SBWA98}, adapted using
Theorem~\ref{thm:ddb_BS} of this paper.%

\begin{theorem}
\label{thm:margin_BS}
Let $\Delta > 0$.
Suppose inputs are drawn independently
according to a distribution $\PP$ be a probability measure on a Banach space $\BB$ of type $p$ and type constant $T$, and $\EE_{\mathcal{P}} \|x\|^p  = r^p < \infty$. If we succeed in
correctly classifying $\ell$ such inputs by a maximum margin hyperplane of margin $\De$, then with confidence $1 - \de$
the generalization error will be bounded from above by
$$ \eps :=  \frac{\tilde{O}\left( \frac{   Tr \ell^{\frac{1}{p}}}{\Delta}\right)+ \log \frac{1}{\de}}{\ell},$$
   where $\tilde{O}$ hides multiplicative polylogarithmic factors involving $\ell, T, r$ and $\De$.
\end{theorem}
Specializing this theorem to a Hilbert space, we have the following theorem 
as a corollary.
\begin{theorem}
\label{thm:margin_HS}
Let $\Delta > 0$.
Suppose inputs are drawn independently
according to a distribution $\PP$ be a probability measure on a Hilbert space $\HH$, and $\EE_{\mathcal{P}} \|x\|^2  = r^2 < \infty$. If we succeed in
correctly classifying $\ell$ such inputs by a maximum margin  hyperplane with margin $\De$, then with confidence $1 - \de$
the generalization error will be bounded from above by
$$ \eps :=  \frac{\tilde{O}\left( \frac{   r \ell^{\frac{1}{2}}}{\Delta}\right)+ \log \frac{1}{\de}}{\ell},$$
   where $\tilde{O}$ hides multiplicative polylogarithmic factors involving $\ell, r$ and $\De$.
\end{theorem}
Note that Theorem~\ref{thm:margin_HS} is an analogue of Theorem~4.17 of the 
journal version of~\cite{SBWA98}, but adapted using Theorem~\ref{thm:ddb_HS} 
of this paper.
In particular, note that this theorem does not assume that the distribution 
is contained in a ball of some radium $R$, but instead it assumes only that 
some moment of the distribution is bounded.

\subsection{Outline of the paper}
\label{sxn:intro_outline}

In the next section, Section~\ref{sxn:background}, we review some technical
preliminaries that we will use in our subsequent analysis.
Then, in Section~\ref{sxn:gapHS}, we state and prove our main result for
gap-tolerant learning in a Hilbert space, and we show how this result can be
used to prove our two main theorems in maximum margin learning.
Then, in Section~\ref{sxn:gapBS}, we state and prove an extension of our
gap-tolerant learning result to the case when the gap is measured with
respect to more general Banach space norms; and
then, in Sections~\ref{sxn:discussion} and~\ref{sxn:conclusion} we
provide a brief discussion and conclusion.
Finally, for completeness, in Appendix~\ref{sxn:learnHT}, we will provide 
a bound for exact (as opposed to maximum margin) learning in the case in 
which the probability that an entry is nonzero (as opposed to the value of 
that entry) decays in a heavy-tailed manner.


\section{Background and preliminaries}
\label{sxn:background}

In this paper, we consider the supervised learning problem of binary
classification, \emph{i.e.}, we consider an input space $\mathcal{X}$
(\emph{e.g.}, a Euclidean space or a Hilbert space) and an output space
$\mathcal{Y}$, where $\mathcal{Y}=\{-1,+1\}$, and where the data consist of
pairs $(X,Y) \in \mathcal{X} \times \mathcal{Y}$ that are random variables
distributed according to an unknown distribution.
We shall assume that for any $X$, there is at most one pair $(X, Y)$ that is
observed.
We observe $\ell$ i.i.d. pairs $(X_i,Y_i), i=1,\dots,\ell$ sampled according
to this unknown distribution, and the goal is to construct a classification
function $\a:\mathcal{X} \rightarrow \mathcal{Y}$ which predicts
$\mathcal{Y}$ from $\mathcal{X}$ with low probability of error.

Whereas an ordinary linear hyperplane classifier consists of an oriented
hyperplane, and points are labeled $\pm1$, depending on which side of the
hyperplane they lie, a \emph{gap-tolerant classifier} consists of an
oriented hyperplane and a margin of thickness $\Delta$ in some norm.
Any point outside the margin is labeled $\pm 1$, depending on which side of
the hyperplane it falls on, and all points within the margin are declared
``correct,'' without receiving a $\pm 1$ label.
This latter setting has been considered in~\cite{Vapnik98,Burges98} (as a
way of implementing structural risk minimization---apply empirical risk
minimization to a succession of problems, and choose where the gap $\Delta$
that gives the minimum risk bound).

The
\emph{risk} $R(\a)$ of a linear hyperplane classifier $\alpha$ is the
probability that $\a$ misclassifies a random data point $(x, y)$ drawn from
$\PP$; more formally, $R(\a) := \E_{\PP}[\a(x) \neq y]$.
Given a set of $\ell$ labeled data points
$(x_1, y_1), \dots, (x_\ell, y_\ell)$, the \emph{empirical risk}
$R_{emp}(\a, \ell)$ of a linear hyperplane classifier $\alpha$ is the
frequency of misclassification on the empirical data; more formally,
$R_{emp}(\a, \ell) := \frac{1}{\ell}\sum_{i=1}^\ell \II[x_i \neq y_i]$, where
$\II[\cdot]$ denotes the indicator of the respective event.
The risk and empirical risk for gap-tolerant classifiers are defined in the
same manner.
Note, in particular, that data points labeled as ``correct'' do not
contribute to the risk for a gap-tolerant classifier, \emph{i.e.}, data
points that are on the ``wrong'' side of the hyperplane but that are within
the $\Delta$ margin are not considered as incorrect and do not contribute
to the risk.

In classification, the ultimate goal is to find a classifier that minimizes
the true risk, \emph{i.e.}, $\arg\min_{\a \in \Lambda} R(\a)$, but since
the true risk $R(\a)$ of a classifier $\a$ is unknown, an empirical
surrogate is often used.
In particular,
\emph{Empirical Risk Minimization (ERM)} is the procedure of choosing a
classifier $\a$ from a set of classifiers $\Lambda$ by minimizing the
empirical risk $\arg\min_{\a \in \Lambda} R_{emp}(\a, \ell)$.
The consistency and rate of convergence of ERM---see~\cite{Vapnik98} for
precise definitions---can be related to uniform bounds on the difference
between the empirical risk and the true risk over all $\a \in \Lambda$.
There is a large body of literature on sufficient conditions for this kind
of uniform convergence.
For instance, the VC dimension is commonly-used toward this end.
Note that, when considering gap-tolerant classifiers, there is an additional
caveat, as one obtains uniform bounds only over those gap-tolerant
classifiers that do not contain any data points in the margin---the
appendix A.2 of~\cite{Burges98} addresses this issue.

In this paper, our main emphasis is on the annealed entropy:

\begin{definition} [Annealed Entropy]
\label{def:an_ent}
Let $\PP$ be a probability measure supported on a vector space $\HH$.
Given a set $\Lambda$ of decision rules and a set of points
$Z = \{z_1, \dots, z_\ell\} \subset \HH$, let $\Nl(z_1, \dots, z_\ell)$ be
the number of ways of labeling   $\{z_1, \dots, z_\ell\} $ into positive and
negative samples such that there exists a gap-tolerant classifier that
predicts {\it incorrectly} the label of each $z_i$.
Given the above notation,
\beqs
\Hnl(k) := \ln \E_{\PP^{\times k}} \Nl(z_1, \dots, z_k)
\eeqs
is the annealed entropy of the classifier $\Lambda$ with respect to $\PP$.
\end{definition}
Note that although we have defined the annealed entropy for general 
decision rules, below we will consider the case that $\Lambda$ consists of 
linear decision rules.

As the following theorem states, the annealed entropy of a classifier can be
used to get an upper bound on the generalization error.
This follows from Theorem $8$ in \cite{Burges98} and a remark on page 198
of~\cite{DevGyoLug96}. Note that the class $\Lambda^*$ is itself random, and consequently, $\sup_{\a \in \Lambda^*}
        {R(\a) - R_{emp}(\a, \ell)}$ is the supremum over a random class.
\begin{theorem} 
\label{thm:Vapnik}
Let $\Lambda^*$ be the family of all gap-tolerant classifiers such that no data point lies inside the margin. Then,
$$
\p\left[\sup_{\a \in \Lambda^*}
        {R(\a) - R_{emp}(\a, \ell)}
        > \epsilon\right]
    <  8 \exp{\left(\left({H_{ann}^\Lambda(\ell)}\right) - \frac{\epsilon^2\ell}{32}\right)}
$$
holds true, for any number of samples $\ell$ and for any error parameter
$\epsilon$.
\end{theorem}
The key property of the function class that leads to uniform bounds is the
sublinearity of the logarithm of the expected value of the ``growth
function,'' which measures the number of distinct ways in which a data set
of a particular size can be split by the function class.
A finite VC bound guarantees this in a distribution-free setting.
The annealed entropy is a distribution-specific measure,
\emph{i.e.}, the same family of classifiers can have different annealed
entropies when measured with respect to different distributions.
For a more detailed exposition of uniform bounds in the context of gap-tolerant classifiers, we refer the reader
to~(\cite{Burges98}, Appendix A.2).

Note also that normed vector spaces (such as Hilbert spaces and Banach
spaces) are relevant to learning theory for the following reason.
Data are often accompanied with an underlying metric which carries
information about how likely it is that two data points have the same label.
This makes concrete the intuition that points with the same class label are
clustered together.
Many algorithms cannot be implemented over an arbitrary metric space, but
require a linear structure.
If the original metric space does not have such a structure, as is the case
when classifying for example, biological data or decision trees, it is
customary to construct a feature space representation, which embeds data
into a vector space.
We will be interested in the commonly-used Hilbert spaces, in which distances
in the feature space are measure with respect to the $\ell_2$ distance (as
well as more general Banach spaces, in Section~\ref{sxn:gapBS}).

Finally, note that our results where the margin is measured in $\ell_2$ can
be transferred to a setting with kernels.
Given a kernel $k(\cdot, \cdot)$, it is well known that linear classification
using a kernel $k(\cdot, \cdot)$ is equivalent to mapping $x$ onto the
functional $k(x, \cdot)$ and then finding a separating halfspace in the
Reproducing Kernel Hilbert Space (RKHS) which is the  Hilbert Space generated
by the functionals of the form $k(x, \cdot)$.
Since the span of any finite set of points in a Hilbert Space can be
isometrically embedded in $\ell_2$, our results hold in the setting of
kernel-based learning as well, when one first uses the feature map
$x \mapsto k(x, \cdot)$ and works in the RKHS.


\section{Gap-tolerant classifiers in Hilbert spaces}
\label{sxn:gapHS}

In this section, we state and prove Theorem~\ref{thm:ddb_HS}, our main result
regarding an upper bound for the annealed entropy of gap-tolerant classifiers
in $\ell_2$.
This result is of independent interest, and it was used in a crucial way in
the proof of Theorems~\ref{thm:learn_heavytail} and~\ref{thm:learn_spectral}.
We start in Section~\ref{sxn:gapHS_annent} with the statement and proof of
Theorem~\ref{thm:ddb_HS}, and then in Section~\ref{sxn:gapHS_vcdim} we bound
the VC dimension of gap-tolerant classifiers over a ball of radius $R$.
Then, in Section~\ref{sxn:gapHS_HTapp}, we apply these results to prove our
main theorem on learning with heavy-tailed data, and finally in
Section~\ref{sxn:gapHS_SKapp}, we apply these results to prove our main
theorem on learning with spectral kernels.

\subsection{Bound on the annealed entropy of gap-tolerant classifiers in Hilbert spaces}
\label{sxn:gapHS_annent}

The following theorem is our main result regarding an upper bound for
the annealed entropy of gap-tolerant classifiers.
The result holds for gap-tolerant classification in a Hilbert space,
\emph{i.e.}, when the distances in the feature space are measured with
respect to the $\ell_2$ norm.
Analogous results hold when distances are measured more generally, as we
will describe in Section~\ref{sxn:gapBS}.

\begin{theorem} [Annealed entropy; Upper bound; Hilbert Space]
\label{thm:ddb_HS}
Let $\PP$ be a probability measure on a Hilbert space $\HH$,
and let $\Delta>0$.
If $\EE_{\mathcal{P}} \|x\|^{2}  = r^{2} < \infty$, then
then the annealed entropy of gap-tolerant classifiers in $\HH$,
where the gap is $\Delta$, is
$$
\Hnl(\ell)
   \leq \left( \ell^{\frac{1}{2}}\left(\frac{r}{\Delta}\right) + 1\right)
        (1+\ln(\ell+1))  .
$$
\end{theorem}
\begin{proof}
Let $\ell$ independent, identically distributed (i.i.d) samples
$z_1, \dots, z_\ell$ be chosen from $\mathcal{P}$.
We partition them into two classes:
$$X = \{x_1, \dots, x_{\ell-k}\} := \{z_i\,\, |\,\, \|z_i\| > R\},$$ and
$$Y = \{y_1, \dots, y_k\} := \{z_i \,\,|\,\, \|z_i\| \leq R\}.$$
Our objective is to bound from above the annealed entropy
$\Hnl(\ell)=\ln\EE[N^\Lambda(z_1, \dots, z_\ell)]$.
By Lemma~\ref{lem:submul}, $N^{\Lambda}$ is sub-multiplicative.
Therefore,
$$N^{\Lambda}(z_1, \dots, z_\ell) \leq N^{\Lambda}(x_1, \dots,
x_{\ell-k}) N^{\Lambda}(y_1, \dots, y_k).$$ Taking an expectation
over $\ell$ i.i.d samples from $\mathcal{P}$,
$$\EE[N^{\Lambda}(z_1, \dots, z_\ell)] \leq \EE[N^{\Lambda}(x_1, \dots, x_{\ell-k})N^{\Lambda}(y_1, \dots,
y_k)].$$
Now applying Lemma~\ref{lem:vap1}, we see that
$$
\EE[N^{\Lambda}(z_1, \dots, z_\ell)]
   \leq \EE[N^{\Lambda}(x_1, \dots, x_{\ell -k})(k+1)^{R^2/\Delta^2+1}]  .
$$
Moving $(k+1)^{R^2/\Delta^2+1}$ outside this expression,
$$
\EE[N^{\Lambda}(z_1, \dots, z_\ell)]
   \leq \EE[N^{\Lambda}(x_1, \dots, x_{\ell -k})](k+1)^{R^2/\Delta^2+1}   .
$$
Note that $N^{\Lambda}(x_1, \dots, x_{\ell-k})$ is always bounded above by
$2^{\ell-k}$ and that the random variables $\I[E_i [\|x_i\| > R]]$ are i.i.d.
Let $\rho = \p[{\|x_i\| > R}]$, and note that
$\ell-k$ is the sum of $\ell$ independent Bernoulli variables.
Moreover, by Markov's inequality,
$$
\p[\|z_i\|>R]  \,\, \leq \,\, \frac{\EE[\|z_i\|^2]}{R^2}  ,
$$
and therefore $\rho \leq (\frac{r}{R})^2$.
In addition,
 $$\EE[N^{\Lambda}(x_1, \dots, x_{\ell-k})] \leq \EE[2^{\ell-k}].$$
 Let $I[\cdot]$ denote an indicator variable. $\EE[2^{\ell-k}]$ can be written as
 $$\prod_{i=1}^\ell \EE[2^{I[\|z_i\|> R]}] = (1+\rho)^\ell \leq
 e^{\rho \ell}.$$
Putting everything together, we see that
\beq
\label{eqn:jensen1}
\EE[N^{\Lambda}(z_1, \dots, z_\ell)]
    \leq
  \exp\left(\ell \left(\frac{r}{R}\right)^2
            + \ln(\ell+1)\left(\frac{R^2}{\Delta^2} + 1\right)\right)   .
\eeq
If we substitute $R = (\ell r^2 \Delta^2)^{\frac{1}{4}}$, it follows that
\begin{eqnarray*}
\Hnl(\ell)
   &=&    \log \EE\left[N^{\Lambda}(z_1, \dots, z_\ell)\right]   \\
   &\leq& \left(\ell^{\frac{1}{2}}\left(\frac{r}{\Delta}\right) + 1\right) (1+\ln(\ell+1))  .
\end{eqnarray*}
\end{proof}

For ease of reference, we note the following easily established fact about
$\Nl$.
This lemma is used in the proof of Theorem~\ref{thm:ddb_HS} above and
Theorem~\ref{thm:ddb_BS} below.

\begin{lemma}
\label{lem:submul}
Let $\{x_1, \dots, x_\ell\} \cup \{ y_1, \dots, y_k\}$ be a partition of the
data $Z$ into two parts.
Then, $\Nl$ is submultiplicative in the following sense:
$$
\Nl(x_1, \dots, x_\ell, y_1, \dots y_k)
   \leq \Nl(x_1, \dots, x_\ell) \Nl(y_1, \dots, y_k)  .
$$
\end{lemma}
\begin{proof}
This holds because any partition of
$Z := \{x_1, \dots, x_\ell, y_1, \dots, y_k\}$
into two parts by an element $\II \in \Lambda$ induces such a partition for
the sets $\{x_1, \dots, x_\ell\}$ and $\{y_1, \dots, y_\ell\}$, and for any
pair of partitions of $\{x_1, \dots, x_\ell\}$ and $\{y_1, \dots, y_k\}$,
there is at most one partition of $Z$ that induces them.
\end{proof}

\subsection{Bound on the VC dimension of gap-tolerant classifiers in Hilbert spaces}
\label{sxn:gapHS_vcdim}

As an intermediate step in the proof of Theorem~\ref{thm:ddb_HS}, we
needed a bound on the VC dimension of a gap-tolerant classifier within a
ball of fixed radius.
Lemma~\ref{lem:vap1} below provides such a bound and is due to
Vapnik~\cite{Vapnik98}.
Note, though, that in the course of his proof (See~\cite{Vapnik98},
page 353.), Vapnik states, without further justification, that due to
symmetry the set of points that is extremal in the sense of being the
hardest to shatter with gap-tolerant classifiers is the regular simplex.
Attention has also been drawn to this fact by Burges (\cite{Burges98},
footnote~20), who mentions that a rigorous proof of this fact seems to be
absent.
Vapnik's claim has since been proved by Hush and
Scovel~\cite{HS01}.
Here, we provide a new proof of Lemma~\ref{lem:vap1}.
It is simpler than previous proofs, and in Section~\ref{sxn:gapBS} we will
see that it generalizes to cases when the margin is measured with norms other
than~$\ell_2$.

\begin{lemma} [VC Dimension; Upper bound; Hilbert Space]
\label{lem:vap1}
In a Hilbert-space,
the VC dimension of a gap-tolerant classifier
whose margin is $\Delta$
over a ball of radius $R$
can by bounded above by
$\lfloor \frac{R^2}{\Delta^2} \rfloor + 1$.
\end{lemma}
\begin{proof}
Suppose the VC dimension is $n$. Then there exists a set of $n$
points $X = \{x_1, \dots, x_n\}$ in $B(R)$ that can be completely
shattered using gap-tolerant classifiers. We will consider  two
cases, first that $n$ is even, and then that $n$ is odd.

First, assume that $n$ is even, i.e., that $n=2k$ for some positive
integer $k$. We apply the probabilistic method to obtain a upper
bound on $n$. Note that for every set $S \subseteq [n]$, the set
$X_S := \{x_i | i \in S\}$ can be separated from $X - X_S$ using a
gap-tolerant classifier. Therefore the distance between the
centroids (respective centers of mass) of these two sets is greater
or equal to $2\Delta$. In particular, for each  $S$ having $k=n/2$
elements,
$$
\|\frac{\sum_{i \in S}x_i}{k} - \frac{\sum_{i \not \in S}x_i}{k}\|
   \geq 2\Delta   .
$$
Suppose now that $S$ is chosen uniformly at random from the ${n
\choose k}$ sets of size $k$. Then,
\begin{eqnarray*}
4 \Delta^2
   &\leq&  \EE\left[\|\frac{\sum_{i \in S}x_i}{k} - \frac{\sum_{i \not \in S}x_i}{k}\|^2\right]   \\
   &=&      k^{-2}\left\{\frac{2k+1}{2k}\sum_{i=1}^n \|x_i\|^2 - \frac{\|\sum_1^n x_i\|^2}{2k} \right\}   \\
   &\leq&  \frac{4(n+1)}{n^2} R^2    .
\end{eqnarray*}
Therefore,
\begin{eqnarray*}
\Delta^2
   &\leq&  \frac{n+1}{n^2} R^2   \\
   &<&  \frac{R^2}{n-1}
\end{eqnarray*}
and so
$$
n < \frac{R^2}{\Delta^2} + 1  .
$$

Next, assume that $n$ is odd. We perform a similar calculation for
$n = 2k+1$. As before, we average over all sets $S$ of cardinality
$k$ the squared distance between the centroid of $X_S$ and the
centroid (center of mass) of  $X-X_S$. Proceeding as before,
\begin{eqnarray*}
4\Delta^2
   & \leq & \EE\left[\|\frac{\sum_{i \in S}x_i}{k} - \frac{\sum_{i \not \in S}x_i}{k+1}\|^2\right]  \\
   & = &    \frac{\sum_{i=1}^n \|x_i\|^2 (1 + \frac{1}{2n}) - \frac{1}{2n}\|\sum_{1 \leq i \leq n} x_i\|^2}{k(k+1)}  \\
   & \leq & \frac{\sum_{i=1}^n \|x_i\|^2 (1 + \frac{1}{2n})}{k(k+1)}  \\
   & = &    \frac{4k+3}{2k(2k+1)(k+1)}\{(2k+1)R^2\}  \\
   & < &    \frac{4R^2}{n-1}  .
\end{eqnarray*}
Therefore, $n < \frac{R^2}{\Delta^2} + 1$.
\end{proof}

\subsection{Learning with heavy-tailed data: proof of Theorem~\ref{thm:learn_heavytail}}
\label{sxn:gapHS_HTapp}

\begin{proof}
For a random data sample $x$,
 \beq \E \|\x\|^2 & \leq & \sum_{i=1}^n
(C i^{-\a})^2 \\ & \leq &  C^2 \zeta(2 \a),\eeq
where $\zeta$ is the Riemann zeta function.
The theorem then follows from Theorem~\ref{thm:margin_HS}.
\end{proof}

\subsection{Learning with spectral kernels: proof of Theorem~\ref{thm:learn_spectral}}
\label{sxn:gapHS_SKapp}

\begin{proof}
A Diffusion Map for the graph $G=(V, E)$ is the feature map that
associates with a vertex $x$, the feature vector $\x = (\l_1^\a
f_1(x), \dots, \l_m^\a f_m(x))$, when the eigenfunctions
corresponding to the top $m$ eigenvalues are chosen. Let $\mu$ be
the uniform distribution on $V$ and $|V| = n$. We note that if the
$f_j$ are normalized eigenfunctions, \emph{i.e.}, $\forall j, \sum_{x \in V}
f_j(x)^2 = 1,$ 
\beq 
\E \|\x\|^2 & = & \frac{\sum_{i=1}^m {\l_i^{2\a}}}{n} \leq \frac{\sum_{i=1}^n {\l_i^{2\a}}}{n}  \leq  1.
\eeq 
The above inequality holds because
the eigenvalues have magnitudes that are less or equal to $1$:
$$1 = \l_1 \geq \dots \geq \l_n \geq -1.$$ 
The theorem then follows from Theorem~\ref{thm:margin_HS}.
\end{proof}


\section{Gap-tolerant classifiers in Banach spaces}
\label{sxn:gapBS}

In this section, we state and prove Theorem~\ref{thm:ddb_BS}, our main result
regarding an upper bound for the annealed entropy of gap-tolerant classifiers
in a Banach space.
We start in Section~\ref{sxn:gapBS_prelim} with some technical preliminaries;
then in Section~\ref{sxn:gapBS_vcdim} we bound the VC dimension of
gap-tolerant classifiers in Banach spaces over a ball of radius $R$; and
finally in Section~\ref{sxn:gapBS_annent} we prove Theorem~\ref{thm:ddb_BS}.
We include this result for completeness since it is of theoretical interest;
since it follows using similar methods to the analogous results for Hilbert
spaces that we presented in Section~\ref{sxn:gapHS}; and since this result
is of potential practical interest in cases where modeling the relationship
between data elements as a dot product in a Hilbert space is too
restrictive, \emph{e.g.}, when the data are extremely sparse and
heavy-tailed.
For recent work in machine learning on Banach spaces,
see~\cite{DerLee,MP04,Men02,ZXZ09}.

\subsection{Technical preliminaries}
\label{sxn:gapBS_prelim}

Recall the definition of a Banach space from Definition~\ref{def:rad} above.
We next state the following form of the Chernoff bound, which we will use in
the proof of Lemma~\ref{lem:banach_ub} below.

\begin{lemma} [Chernoff Bound]
\label{lem:chernoff}
Let $X_1, \dots, X_n$ be discrete independent random variables such that
$\EE[X_i]=0$ for all $i$ and $|X_i| \leq 1$ for all $i$.
Let $X = \sum_{i=1}^n X_i$ for all $i$ and $\sigma^2$ be the variance of $X$.
Then
$$
\p[|X| \geq \lambda \sigma] \leq 2e^{-\lambda^2/4}
$$
for any $0 \leq \lambda \leq 2\sigma$.
\end{lemma}

\subsection{Bounds on the VC dimension of gap-tolerant classifiers in Banach spaces}
\label{sxn:gapBS_vcdim}

The idea underlying our new proof of Lemma~\ref{lem:vap1} (of
Section~\ref{sxn:gapHS_vcdim}, and that provides an upper bound on the
VC dimension of a gap-tolerant classifier in Hilbert spaces) generalizes to
the case when the the gap is measured in more general Banach spaces.
We state the following lemma for a Banach space of type $p$ with type
constant $T$.
Recall, \emph{e.g.}, that $\ell_p$ for $p \geq 1$ is a Banach space of type
$\min(2, p)$ and type constant $1$.

\begin{lemma} [VC Dimension; Upper bound; Banach Space]
\label{lem:banach_ub}
In a Banach Space of type $p$ and type constant $T$,
the VC dimension of a gap-tolerant classifier
whose margin is $\Delta$
over a ball of radius $R$
can by bounded above by
$ \left(\frac{3T R}{\Delta}\right)^{\frac{p}{p-1}} + 64$
\end{lemma}
\begin{proof}
Since a general Banach space does not possess an inner product, the proof
of Lemma~\ref{lem:vap1} needs to be modified here.
To circumvent this difficulty, we use Inequality~(\ref{ineq:rad})
determining the Rademacher type of $\mathcal{B}$.
This, while permitting greater generality, provides weaker bounds than
previously obtained in the Euclidean case.
Note that if $\mu :=\frac{1}{n}\sum_{i=1}^n x_i$, then by repeated application of the
Triangle Inequality,
\begin{eqnarray*}
  \|x_i -
\mu\| & \leq  & (1-\frac{1}{n})\|x_i\| + \sum_{j\neq i} \frac{\|x_j\|}{n}\\
& < & 2 \sup_i \|x_i\|.
\end{eqnarray*}
This shows that if we start with $x_1, \dots, x_n$ having norm $\leq
R$, $\|x_i - \mu\| \leq 2R$ for all $i$. The property of being
shattered by gap-tolerant classifiers is translation invariant.
Then, for $ \emptyset \subsetneq S \subsetneq [n]$, it can be
verified that
\begin{eqnarray}
\nonumber
2\Delta
   &\leq& \left\| \frac{\sum_{i \in S} (x_i-\mu)}{|S|} - \frac{\sum_{i \not\in S} (x_i-\mu)}{n - |S|} \right\|  \\
   &=& \frac{n}{2|S|(n-|S|)}\left\|\sum_{i \in S} (x_i -\mu) - \sum_{i \not\in S} (x_i - \mu)\right\|   .
\label{eleven}
\end{eqnarray}
The Rademacher Inequality states that
\beq\label{ineq:rad}\EE_\epsilon[\|\sum_{i=1}^n \epsilon_i x_i\|^p]
\leq T^p \sum_{i=1}^n \|x_i\|^p.\eeq Using the version of Chernoff's
bound in Lemma~\ref{lem:chernoff} \beq \label{twelve}
\p[|\sum_{i=1}^n \epsilon_i| \leq \lambda \sqrt{n}] \geq 1-  2
e^{-\lambda^2/4}. \eeq We shall denote the above event by
$E_{\lambda}$. Now, let $x_1, \dots, x_n$ be $n$ points in
$\mathcal{B}$ with a norm less or equal to $R$. Let $\mu =
\frac{\sum_{i=1}^n x_i}{n}$ as before.
\begin{eqnarray*}
 2^pT^p n R^p & \geq &  2^pT^p \sum_{i=1}^n \|x_i\|^p\\
                    & \geq & T^p \sum_{i=1}^n \|x_i - \mu\|^p\\
                    & \geq & \EE_\epsilon [\|\epsilon_i (x_i-\mu)\|^p]\\
                    & \geq & \EE_\epsilon [\|\epsilon_i (x_i-\mu)\|^p| E_{\lambda}]\,\,\p[E_{\lambda}]\\
                    & \geq & \EE_\epsilon [(n - \lambda^2)^p(2\Delta)^p (1 - 2
                    e^{-\lambda^2/4})]\,\,
\end{eqnarray*}
The last inequality follows from (\ref{eleven}) and (\ref{twelve}).
We infer from the preceding sequence of inequalities that
$$n^{p-1} \leq 2^p T^p \left(\frac{R}{\Delta}\right)^p
\left\{(1-\frac{\lambda^2}{n})^p
(1-2e^{-\lambda^2/4})\right\}^{-1}.$$ The above is true for any
$\lambda \in (0, 2\sqrt{n})$, by the conditions in the Chernoff
bound stated in Lemma~\ref{lem:chernoff}. If $n \geq 64$, choosing
$\lambda$ equal to $8$ gives us $n^{p-1} \leq 3^p T^p
\left(\frac{R}{\Delta}\right)^p.$ Therefore, it is always true that
$n \leq \left(\frac{3TR}{\Delta}\right)^\frac{p}{p-1} + 64.$
\end{proof}

Finally, for completeness, we next state a lower bound for VC dimension of
gap-tolerant classifiers when the margin is measured in a norm that is
associated with a Banach space of type $p \in (1, 2]$.
Since we are interested only in a lower bound, we consider the special case
of $\ell_p^n$.
Note that this argument does not immediately generalize to Banach spaces of
higher type because for $p >2$, $\ell_p$ has type $2$.

\begin{lemma} [VC Dimension; Lower Bound; Banach Space]
\label{lem:banach_lb}
For each $p \in (1, 2]$, there
exists a Banach space of type $p$ such that the VC dimension of gap-tolerant
classifiers with gap $\Delta$ over a ball of radius $R$ is
greater or equal to
$$\left(\frac{R}{\Delta}\right)^{\frac{p}{p-1}}.$$ Further, this
bound is achieved when the space is $\ell_p$.
\end{lemma}
\begin{proof}
We shall show that the first $n$ unit norm basis vectors in the
canonical basis can be shattered using gap-tolerant classifiers,
where $\Delta = n^{\frac{1-p}{p}}$. Therefore in this case, the VC
dimension is $\geq (\frac{R}{\Delta})^\frac{p}{p-1}$. Let $e_j$ be
the $j^{th}$ basis vector. In order to prove that the set $\{e_1,
\dots, e_n\}$ is shattered, due to symmetry under permutations, it
suffices to prove that for each $k$, $\{e_1, \dots, e_k\}$ can be
separated from $\{e_{k+1}, \dots, e_{n}\}$ using a gap-tolerant
classifier. Points in $\ell_p$ are infinite sequences $(x_1, \dots
)$ of finite $\ell_p$ norm. Consider the hyperplane $H$ defined by
$\sum_{i=1}^k x_i - \sum_{i=k+1}^n x_i = 0$. Clearly, it separates
the sets in question. We may assume $e_j$ to be $e_1$, replacing if
necessary, $k$ by $n-k$. Let $x = \inf_{y \in H} \|e_1 - y\|_p.$
Clearly, all coordinates $x_{n+1}, \dots $ of $x$ are $0$. In order
to get a lower bound on the $\ell_p$
distance, we use the power-mean inequality:
If $p \geq 1$, and $x_1, \dots, x_n \in \mathbb{R}$,
$$\left(\frac{\sum_{i=1}^n |x_i|^p}{n}\right)^{\frac{1}{p}} \geq
\frac{\sum _{i=1}^n |x_i|}{n}.$$ This implies that
\begin{eqnarray*}
\|e_1 -x\|_p
   & \geq & n^{\frac{1-p}{p}} \|e_1 - x\|_1  \\
   & = &    n^{\frac{1-p}{p}}\left(|1-x_1| + \sum_{i=2}^n |x_i|\right)  \\
   & \geq & n^{\frac{1-p}{p}}\left(1 - \sum_{i=1}^k x_i + \sum_{i=k+1}^n x_i\right)  \\
   & = &    n^{\frac{1-p}{p}}  .
\end{eqnarray*}
For $p>2$, the type of $\ell_p$ is $2$~\cite{LedouxTalagrand}. Since
$\frac{p}{p-1}$ is a decreasing function of $p$ in this regime,  we
do not recover any useful bounds.
\end{proof}

\subsection{Bound on the annealed entropy of gap-tolerant classifiers in Banach spaces}
\label{sxn:gapBS_annent}

The following theorem is our main result regarding an upper bound for
the annealed entropy of gap-tolerant classifiers in Banach spaces.
Note that the $\ell_2$ bound provided by this theorem is slightly weaker
than that provided by Theorem~\ref{thm:ddb_HS}.
Note also that it may seem counter-intuitive that in the case of $\ell_2$
(\emph{i.e.}, when we set $\gamma = 2$), the dependence of $\Delta$ is
$\Delta^{-1}$, which is weaker than in the VC bound, where it is
$\Delta^{-2}$.
The explanation is that the bound on annealed entropy here depends on the
number of samples $\ell$, while the VC dimension does not.
Therefore, the weaker dependence on $\Delta$ is compensated for by a term
that in fact tends to $\infty$ as the number of samples~$\ell \ra \infty$.

\begin{theorem} [Annealed entropy; Upper bound; Banach Space]
\label{thm:ddb_BS}
Let $\PP$ be a probability measure on a Banach space $\BB$ of type $p$ and
type constant $T$.
Let $\gamma, \Delta > 0$, and let $\eta = \frac{p}{p + \gamma(p-1)}$.
If $\EE_{\mathcal{P}} \|x\|^\gamma  = r^\gamma< \infty$, then
the annealed entropy of gap-tolerant classifiers in $\BB$,
where the gap is $\Delta$, is
$$
\Hnl(\ell)
    \leq
   \left(\eta^{-\eta}(1-\eta)^{-1+\eta} \left(\frac{\ell}{\ln(\ell+1)}
   \left(\frac{3Tr}{\Delta}\right)^\gamma\right)^\eta + 64\right)
   \ln(\ell+1)   .
$$
\end{theorem}
\begin{proof}
The proof of this theorem parallels that of Theorem~\ref{thm:ddb_HS}, except
that here we use Lemma~\ref{lem:banach_ub} instead of Lemma~\ref{lem:vap1}.
We include the full proof for completeness.
Let $\ell$ independent, identically distributed (i.i.d) samples
$z_1, \dots, z_\ell$ be chosen from $\mathcal{P}$.
We partition them into two classes:
$$X = \{x_1, \dots, x_{\ell-k}\} := \{z_i\,\, |\,\, \|z_i\| > R\},$$ and
$$Y = \{y_1, \dots, y_k\} := \{z_i \,\,|\,\, \|z_i\| \leq R\}.$$
Our objective is to bound from above the annealed entropy
$\Hnl(\ell)=\ln\EE[N^\Lambda(z_1, \dots, z_\ell)]$.
By Lemma~\ref{lem:submul}, $N^{\Lambda}$ is sub-multiplicative.
Therefore,
$$N^{\Lambda}(z_1, \dots, z_\ell) \leq N^{\Lambda}(x_1, \dots,
x_{\ell-k}) N^{\Lambda}(y_1, \dots, y_k).$$ Taking an expectation
over $\ell$ i.i.d samples from $\mathcal{P}$,
$$\EE[N^{\Lambda}(z_1, \dots, z_\ell)] \leq \EE[N^{\Lambda}(x_1, \dots, x_{\ell-k})N^{\Lambda}(y_1, \dots,
y_k)].$$
Now applying Lemma~\ref{lem:banach_ub}, we see that
$$
\EE[N^{\Lambda}(z_1, \dots, z_\ell)]
   \leq \EE[N^{\Lambda}(x_1, \dots, x_{\ell -k})(k+1)^{(3TR/\Delta)^{\frac{p}{p-1}}+ 64}]  .
$$
Moving $(k+1)^{((2+o(1)TR/\Delta)^{\frac{p}{p-1}})}$ outside this expression,
$$
\EE[N^{\Lambda}(z_1, \dots, z_\ell)]
   \leq \EE[N^{\Lambda}(x_1, \dots, x_{\ell -k})](k+1)^{(3TR/\Delta)^{\frac{p}{p-1}}+ 64}   .
$$
Note that $N^{\Lambda}(x_1, \dots, x_{\ell-k})$ is always bounded above by
$2^{\ell-k}$ and that the random variables $\I[E_i [\|x_i\| > R]]$ are i.i.d.
Let $\rho = \p[{\|x_i\| > R}]$, and note that
$\ell-k$ is the sum of $\ell$ independent Bernoulli variables.
Moreover, by Markov's inequality,
$$
\p[\|z_i\|>R]  \,\, \leq \,\, \frac{\EE[\|z_i\|^\gamma]}{R^\gamma}  ,
$$
and therefore $\rho \leq (\frac{r}{R})^\gamma$.
In addition,
 $$\EE[N^{\Lambda}(x_1, \dots, x_{\ell-k})] \leq \EE[2^{\ell-k}].$$
 Let $I[\cdot]$ denote an indicator variable. $\EE[2^{\ell-k}]$ can be written as
 $$\prod_{i=1}^\ell \EE[2^{I[\|z_i\|> R]}] = (1+\rho)^\ell \leq
 e^{\rho \ell}.$$
Putting everything together, we see that
\beq
\label{eqn:jensen2}
\EE[N^{\Lambda}(z_1, \dots, z_\ell)]
   \leq \exp\left(\ell \left(\frac{r}{R}\right)^{\gamma} + \ln(k+1)\left(64 + \frac{3T R}{\Delta}\right)^{\frac{p}{p-1}}\right)   .
\eeq
By setting  $\eta :=
\frac{p}{\gamma(p-1) + p},$ and adjusting $R$ so that  $$\ell
\left(\frac{r}{R}\right)^\gamma \eta^{-1} = (1-
\eta)^{-1}\ln(\ell+1)\left(\frac{3TR}{\Delta}\right)^{\frac{p}{p-1}}.$$
We see that
\begin{eqnarray*}
\ell \left(\frac{r}{R}\right)^\gamma  +
\left(\frac{3TR}{\Delta}\right)^{\frac{p}{p-1}}
  &=&  \left(\ell \left(\frac{r}{R}\right)^\gamma \eta^{-1}\right)^\eta
         \left((1- \eta)^{-1}\ln(\ell+1)
            \left(\frac{3TR}{\Delta}\right)^{\frac{p}{p-1}}\right)^{1-\eta} \\
  &=& \eta^{-\eta}(1-\eta)^{-1+\eta}
          \left(\ell\left(\frac{3Tr}{\Delta}\right)^\gamma\right)^\eta  .
\end{eqnarray*}
Thus, it follows that
\begin{eqnarray*}
\Hnl(\ell)
  &=&    \log \EE\left[N^{\Lambda}(z_1, \dots, z_\ell)\right]  \\
  &\leq& \left(\eta^{-\eta}(1-\eta)^{-1+\eta}
            \left(\frac{\ell}{\ln(\ell+1)}
            \left(\frac{3Tr}{\Delta}\right)^\gamma\right)^\eta + 64\right)
            \ln(\ell+1)  .
\end{eqnarray*}
\end{proof}

\section{Discussion}
\label{sxn:discussion}

In recent years, there has been a considerable amount of somewhat-related
technical work in a variety of settings in machine learning.
Thus, in this section we will briefly describe some of the more technical
components of our results in light of the existing related literature.
\begin{itemize}
\item
Techniques based on the use of Rademacher inequalities allow one to obtain
bounds without any assumption on the input distribution as long as the
feature maps are uniformly bounded.
See, \emph{e.g.},~\cite{Gurvits,KP00,BartMen02,Kol01}.
Viewed from this perspective, our results are interesting because the
uniform boundedness assumption is not satisfied in either of the two
settings we consider, although those settings are ubiquitous in
applications.
In the case of heavy-tailed data, the uniform boundedness assumption is not
satisfied due to the slow decay of the tail and the large variability of
the associated features.
In the case of spectral learning, uniform boundedness assumption is not
satisfied since for arbitrary graphs one can have localization and thus
large variability in the entries of the eigenvectors defining the feature
maps.
In both case, existing techniques based on Rademacher inequalities or
VC dimensions fail to give interesting results, but we show that
dimension-independent bounds can be achieved by bounding the annealed
entropy.
\item
A great deal of work has focused on using diffusion-based and spectral-based
methods for nonlinear dimensionality reduction and the learning a nonlinear
manifold from which the data are assumed to be drawn~\cite{SWHSL06}.
These results are very different from the type of learning bounds we
consider here.
For instance, most of those learning results involve convergence to an
hypothesized manifold Laplacian and not of learning process itself, which
is what we consider here.
\item
Work by Bousquet and Elisseeff~\cite{BE02} has focused on establishing
generalization bounds based on stability.
It is important to note that their results assume a given algorithm and
show how the generalization error changes when the data are changed, so
they get generalization results for a given algorithm.
Our results make no such assumptions about working with a given algorithm.
\item
Gurvits~\cite{Gurvits} has used Rademacher complexities to prove upper bounds
for the sample complexity of learning bounded linear functionals on
$\ell_p$ balls.
The results in that paper can be used to derive an upper bound on the VC
dimension of gap-tolerant classifiers with margin $\Delta$ in a ball of
radius $R$ in a Banach space of Rademacher type $p \in (1, 2]$.
Constants were not computed in that paper, therefore our results do not
follow.
Moreover, our paper contains results on distribution specific bounds which
were not considered there.
Finally, our paper considers the application of these tools to the
practically-important settings of spectral kernels and heavy-tailed data
that were not considered there.
\end{itemize}

\section{Conclusion}
\label{sxn:conclusion}

We have considered two simple machine learning problems motivated by recent
work in large-scale data analysis, and we have shown that although
traditional distribution-independent methods based on the VC-dimension fail
to yield nontrivial sampling complexity bounds, we can use
distribution-dependent methods to obtain dimension-independent learning
bounds.
In both cases, we take advantage of the fact that, although there may be
individual data points that are ``outlying,'' in aggregate their effect is
not too large.
Due to the increased popularity of vector space-based methods (as opposed
to more purely combinatorial methods) in machine learning in recent years,
coupled with the continued generation of noisy and poorly-structured data,
the tools we have introduced are likely promising more generally for
understanding the effect of noise and noisy data on popular machine
learning tasks.

\appendix
\section{Exact learning with heavy-tailed data}
\label{sxn:learnHT}

In this appendix section, we state and prove a second result for 
dimension-independent learning from data in which the feature map exhibits 
a heavy-tailed decay.
The heavy-tailed model we consider here is different than that considered 
in Theorem~\ref{thm:learn_heavytail}, and thus we are able to prove bounds 
for exact (as opposed to maximum margin) learning.
Nevertheless, the techniques are similar, and thus we include this result
in this paper for completeness.

Consider the following toy model for classifying web pages using keywords.
One approach to this problem could be to associate with each web page the
indicator vector corresponding to all keywords that it contains.
The dimension of this feature space is the number of possible keywords,
which is typically very large, and empirical evidence indicates that the
frequency of words decays in a heavy-tailed manner.
Thus the VC dimension of the feature space is very large, and in a
distribution-free setting it is not possible to classify data in such a
feature space unless the number of samples is of the order of the VC
dimension.
More generally, one might be interested in a bipartite graph, \emph{e.g.}, an
``advertiser-keyword'' or ``author-to-paper'' graph, in which the nodes are
the stated entities and the edges represent some sort of ``interaction''
between the entities, in which case similar issues arise.

Here, 
we show that if the probability that the $i^{th}$ keyword in the above toy
example is present is heavy-tailed as a function of $i$, then the sample
complexity of the binary classification problem is dimension-independent.
More precisely, the following theorem provides a dimension-independent
(\emph{i.e.}, independent of the size $n$ of the graph and the dimension of
the feature space) upper bound on the number of samples needed to learn by
ERM, with a given accuracy and confidence, a linear hyperplane that
classifies heavy-tailed data into positive and negative labels, under the
assumption that the probability of the $i^{th}$ coordinate of a random data
point being non-zero is less than $C i^{-\a}$ for some $C>0, \a > 1$.
The proof of this result proceeds by providing providing a
dimension-independent upper bound on the annealed entropy of the class of
linear classifiers in $\R^d$, and then appealing to Theorem~\ref{thm:Vapnik}
relating the annealed entropy to the generalization error.

\textbf{Remark:}
Note that although the generalization bound provided by the following
theorem seems to be pessimistic in $\alpha$, the dependence on $\alpha$ is
tight, at least as $\alpha$ tends to $1$.
Clearly, when $\alpha = 1$, the expected number of $1$'s in a random sample
becomes asymptotically equal to $\log n$, where $n$ is the dimension, in
which case, we do not expect a sample complexity that is
dimension-independent.

\begin{theorem} [Bounds for Heavy-Tailed Data]
\label{thm:learnHT_exact}
Let $\PP$ be a probability distribution in $\R^d$.
Suppose $\PP[x_i \neq 0 ] \leq Ci^{-\alpha}$ for some absolute constant
$C>0$, with $\alpha > 1$.
Then, the annealed entropy of ordinary linear hyperplane classifiers is
\beq
\Hnl(\ell)
   & \leq & \left(\frac{C}{\a - 1}\ell^{\frac{1}{\alpha}}+1\right)\ln \ell
\eeq
Consequently, the minimum number of random samples $\ell=\ell(\eps,\delta)$
needed to learn, by ERM, a classifier whose risk differs from the minimum
risk $R(\a)$ by $< \eps \sqrt{R(\a)} $ with probability $> 1-\delta$ is
less than or equal to
\beqs
2\left(\frac{4}{\eps^2}\left(\frac{C2^{\frac{1}{\alpha}}}{\alpha -1}
+ \ln\frac{4}{\delta}\right)\right)^\frac{\a}{\a-1}\ln
\left(\left(\frac{4}{\eps^2}\left(\frac{C2^{\frac{1}{\alpha}}}{\alpha
-1} + \ln\frac{4}{\delta}\right)\right)^\frac{\a}{\a-1}\right)   .
\eeqs
\end{theorem}
\begin{proof}
Let the event that a sample $z_i = (z_{i1}, z_{i2}, \dots)$ has a
non-zero coordinate $z_{ik'}$ for some $k' > \ell^{1/\a}$ be denoted
$E_{i}$. The probability of this event can be bounded as follows. If
$\alpha \neq 1$ and~$k = \ell^{1/\a}$, then \beqs
\p[E_{i}] & = & \p[\exists k' > \ell^{1/\a}, \text{ such that } z_{ik'} \neq 0]\\ & \leq & C\sum_{i=k+1}^\infty i^{-\alpha}\\
                                                            & \leq & \frac{C k^{-\alpha +
                                                            1}}{\a
                                                            -1}   .
\eeqs We partition the $z_i$  into two classes : $$X = \{x_1, \dots,
x_{\ell-m}\} := \{z_i \text{ such that } E_{i}\text{ holds }\}$$ and
$$Y = \{y_1, \dots, y_m\} := \{z_i \text{ such that } E_{i} \text{ does not hold }\}.$$
$\Nl$ is sub-multiplicative by Lemma~\ref{lem:submul}. Taking an
expectation over $\ell$ i.i.d samples from $\PP$, \beqs \EE[\Nl(z_1,
\dots, z_\ell)] & \leq & \EE[\Nl(x_1, \dots, x_{\ell-m})\Nl(y_1,
\dots, y_m)]\eeqs The dimension of the span of $\{y_1, \dots, y_m\}$
is at most $k$, and by a result from VC theory (\cite{Vapnik98}, page 159)
we have $$\Nl(y_1, \dots, y_{m}) \leq \exp(k \ln(\frac{m}{k}) + 1).$$
 Then,
$$\EE[\Nl(z_1, \dots, z_\ell)] \leq \EE[\Nl(x_1, \dots, x_{\ell
-m})e m^k].$$
Moving $e m^k$ outside this expression,
$$
\EE[\Nl(z_1, \dots, z_\ell)]
   \leq \EE[\Nl(x_1, \dots, x_{\ell -k})]e m^k   .
$$
Note that $\Nl(x_1, \dots, x_{\ell-k})$ is always bounded above by
$2^{\ell-k}$ and that the events $E_{1}, E_{2}, \dots$ are independent
identically distributed.
Let $p = \p[E_{i}]$, and note that
$\ell-k$ is the sum of $\ell$ independent $p$-Bernoulli variables.
In addition,
 $$\EE[\Nl(x_1, \dots, x_{\ell-k})] \leq \EE[2^{\ell-k}],$$ and
  $\EE[2^{\ell-k}]$ can be written as
 \beq \prod_{i=1}^\ell (1 + \p[E_i]) & = & (1+p)^\ell \\
 & \leq & e^{p\ell}\\
 & = & e^{\ell\,(\frac{C \, k^{-\a + 1}}{\a - 1})}.\eeq
Putting everything together, we see that
$$
\EE[\Nl(z_1, \dots, z_\ell)]
   \leq e(\ell)^k e^{\frac{C \ell\, k^{-a + 1}}{\a - 1}}   .
$$
Since $k = \ell^{\frac{1}{\alpha}}$, we see that
\beq
\Hnl(\ell)
   & = & \ln \EE[\Nl(z_1, \dots, z_\ell)]  \\
   & \leq & \left(\frac{C}{\a - 1} \ell^{\frac{1}{\alpha}}+1\right)\,\ln (\ell).
\eeq
In order to obtain sample complexity bounds, we need to apply
Theorem~\ref{thm:Vapnik} and substitute the above expression for annealed
entropy.
For the probability that the error of ERM
exceeds $\eps \sqrt{R(\a)}$ to be less than $\delta$ (where $\a$ is
the optimal classifier), it is sufficient that $\ell$  satisfy
$$4 \exp\left(\frac{C 2^{1/\a}}{\a -1} \ell^{\frac{1-\a}{\a}} \ln(2\ell) -
\eps^2/4\right)\ell \leq \delta.$$ For this to be true, it is enough
that
$$\frac{\eps^2 \ell^{1-\frac{1}{\a}} }{4} \geq \frac{C
2^{\frac{1}{\a}}\ln(2\ell)}{\a-1} + \ln(4/\delta).$$ A calculation
shows that
$$\frac{2\alpha \left(\frac{4}{\eps^2}\left(\frac{C2^{\frac{1}{\alpha}}}{\alpha -1}
+ \ln\frac{4}{\delta}\right)\right)^\frac{\a}{\a-1}\ln
\left(\frac{4}{\eps^2}\left(\frac{C2^{\frac{1}{\alpha}}}{\alpha
-1} + \ln\frac{4}{\delta}\right)\right)}{\a -1}$$ is
a value of $\ell$ that satisfies the previous expression.
\end{proof}

\end{document}